\documentclass{article}

\usepackage{microtype}
\usepackage{graphicx}
\usepackage{subfigure}
\usepackage{subcaption}
\usepackage{booktabs}

\usepackage{hyperref}

\usepackage[accepted]{icml2025}

\usepackage{amsmath}
\usepackage{amssymb}
\usepackage{mathtools}
\usepackage{amsthm}
\usepackage{pgfplots}

\usepackage[capitalize,noabbrev]{cleveref}

\theoremstyle{plain}

\theoremstyle{definition}

\theoremstyle{remark}

\usepackage[textsize=tiny]{todonotes}

\icmltitlerunning{Submission and Formatting Instructions for ICML 2025}

\begin{document}

\twocolumn[
\icmltitle{GDO: Gradual Domain Osmosis}



\icmlsetsymbol{equal}{*}

\begin{icmlauthorlist}
\icmlauthor{Zixi Wang}{equal,sch1,sch2}
\icmlauthor{Yubo Huang}{equal,sch1}
\end{icmlauthorlist}

\icmlaffiliation{sch1}{University of Electronic Science and Technology of China, Sichuan, China}
\icmlaffiliation{sch2}{Southwest Jiaotong University, Sichuan, China}

\icmlcorrespondingauthor{Zixi Wang}{zixi-wang@outlook.com}
\icmlcorrespondingauthor{Yubo Huang}{ybforever@my.swjtu.edu.cn}

\icmlkeywords{Machine Learning, ICML}

\vskip 0.3in
]



\printAffiliationsAndNotice{\icmlEqualContribution} 

\begin{abstract}

In this paper, we propose a new method called Gradual Domain Osmosis, which aims to solve the problem of smooth knowledge migration from source domain to target domain in Gradual Domain Adaptation (GDA). Traditional Gradual Domain Adaptation methods mitigate domain bias by introducing intermediate domains and self-training strategies, but often face the challenges of inefficient knowledge migration or missing data in intermediate domains. In this paper, we design an optimisation framework based on the hyperparameter $\lambda$ by dynamically balancing the loss weights of the source and target domains, which enables the model to progressively adjust the strength of knowledge migration ($\lambda$ incrementing from 0 to 1) during the training process, thus achieving cross-domain generalisation more efficiently. Specifically, the method incorporates self-training to generate pseudo-labels and iteratively updates the model by minimising a weighted loss function to ensure stability and robustness during progressive adaptation in the intermediate domain. The experimental part validates the effectiveness of the method on rotated MNIST, colour-shifted MNIST, portrait dataset and forest cover type dataset, and the results show that it outperforms existing baseline methods. The paper further analyses the impact of the dynamic tuning strategy of the hyperparameter $\lambda$ on the performance through ablation experiments, confirming the advantages of progressive domain penetration in mitigating the domain bias and enhancing the model generalisation capability. The study provides a theoretical support and practical framework for asymptotic domain adaptation and expands its application potential in dynamic environments.

\end{abstract}

\section{Introduction}

\begin{figure}
    \centering
    \includegraphics[width=0.5\linewidth]{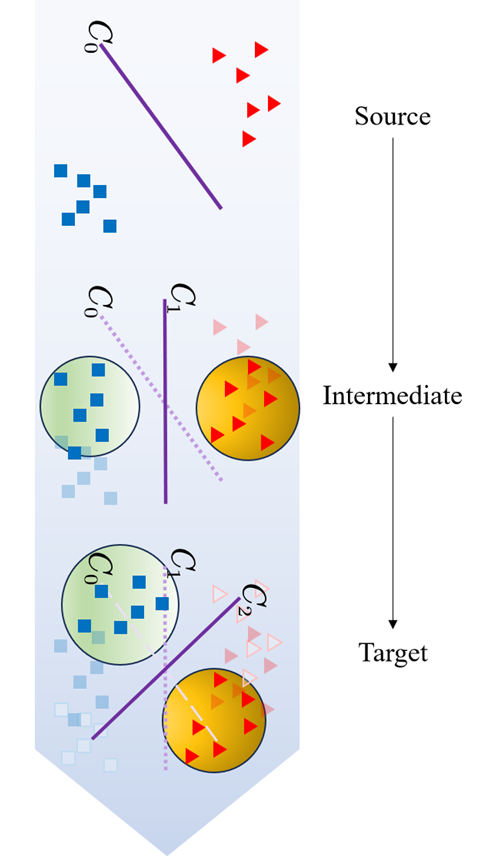}
    \caption{An example of domain shift. The red is source domain and the blue is target domain. The yellow regions are assumed to have zero support in the target distribution.}
    \label{fig1}
\end{figure}

One of the core challenges in machine learning is the domain shift, which occurs when the distribution of data in the training domain (source domain) differs significantly from the distribution in the target domain \cite{ovadia2019can, pan2009survey}, as shown in Figure \ref{fig1}. This problem is prevalent in real-world applications where labeled data in the target domain is scarce or expensive to obtain. Domain adaptation (DA) \cite{li2016structured, zhao2020review} has become an essential strategy for transferring knowledge across domains, particularly in situations where data distributions between the source and target domains are misaligned. Unsupervised domain adaptation (UDA) aims to address this issue by utilizing unlabeled data from the target domain to enhance model performance \cite{liu2022deep}.  However, UDA struggles when domain shifts are large, or when intermediate domain data is minimal or nonexistent.

To address these limitations, recent advances have introduced Gradual Domain Adaptation (GDA), a framework that aims to reduce the drastic shift between domains by employing a more controlled and stepwise knowledge transfer process. \cite{farshchian2018adversarial, kumar2020understanding}. \cite{kumar2020understanding} explored this gradual approach and proposed strategies to improve model stability during adaptation, demonstrating that such a progressive transfer process can significantly mitigate the adverse effects of domain shifts, but challenges remain in efficiently transitioning between domains and maintaining model performance. \cite{chen2021gradual} further refined the GDA framework by introducing new techniques for generating intermediate domains and enhancing the smoothness of the adaptation process. Their approach shows promising results in various tasks, including image classification and object detection, where large domain shifts are prevalent. However, the migration of knowledge across intermediate domains is still suboptimal, leading to slower convergence times and potential overfitting issues \cite{he2023gradual}. In addition, \cite{marsden2024introducing} examined the risks of instability when adapting models through numerous intermediate domains, highlighting the need for novel techniques to control the adaptation process more precisely.

In this work, we aim to address the limitations of GDA by focusing on the efficient migration of knowledge between domains, reducing the risk of instability, and exploring novel methods to improve the effectiveness of the gradual adaptation process. First, we propose a new optimization objective that balances source and target losses dynamically. Secondly, we introduce a self-training mechanism that iteratively refines the model using pseudo-labeled data. Our contributions are as follows:

\begin{itemize}
    \item We designed an optimization framework, namely Gradual Domain Osmosis (GDO), based on the hyperparameter $\lambda$ to achieve more efficient knowledge transfer by dynamically adjusting the weights of the loss functions in the source and target domains. As the training process progresses, the $\lambda$ value gradually increases from 0 to 1, allowing the model to gradually adjust the strength of knowledge transfer, thereby more effectively achieving cross-domain generalization.
    \item In order to ensure stability and robustness during the gradual adaptation process in the intermediate domain, the GDO method combines self-training technology to generate pseudo labels and iteratively updates the model by minimizing the weighted loss function. This method helps to alleviate the performance degradation caused by domain shift and improve the generalization ability of the model between different domains.
    \item The paper verifies the effectiveness of the proposed method through experiments on Rotated MNIST, color-shifted MNIST, Portrait datasets, and Cover type datasets The results show that this method outperforms existing baseline methods.
\end{itemize}

\section{Related Work}

\textbf{Domain Generalization} Domain Generalization (DG) aims to train models that generalize well to unseen target domains without accessing target data during training \cite{muandet2013domain, gluon2017domain}. Unlike UDA, DG does not assume any availability of target domain data, making it highly applicable in scenarios where target environments are unknown or continuously evolving. Strategies in DG include learning robust feature representations through data augmentation, meta-learning, and enforcing invariance to domain-specific variations \cite{li2017domain, balaji2018improving, jin2020meta}. Recent advancements have focused on leveraging meta-learning frameworks to simulate domain shifts during training, thereby preparing models for unseen domains \cite{wang2022understanding}.

\textbf{Unsupervised Domain Adaptation} Unsupervised Domain Adaptation (UDA) tackles the challenge of domain shifts—discrepancies between source (training) and target (testing) domains—by leveraging only unlabeled data from the target domain \cite{farahani2021brief, pmlr-v37-ganin15, Tzeng_2017_CVPR}. The goal is to learn robust, domain-invariant features that enable a model to generalize despite distribution mismatches \cite{pan2009survey, hoffman2018cycada}. Common approaches include aligning source and target feature distributions via measures like maximum mean discrepancy \cite{8767033, Yan_2017_CVPR} or through adversarial learning frameworks \cite{Zhang_2018_CVPR, Volpi_2018_CVPR}. However, recent studies \cite{Kang_2019_CVPR, tang2020discriminative, yang2020mind, zhao2019learning, kumar2020understanding} reveal that naive alignment can introduce misalignment and fail when the shift is severe or the target data present novel variations outside the source’s coverage.

\textbf{Gradual Domain Adaptation} Gradual Domain Adaptation (GDA) addresses scenarios where data shift gradually rather than abruptly, allowing the overall shift to be decomposed into a sequence of smaller steps \cite{farshchian2018adversarial, kumar2020understanding}. By introducing intermediate domains and employing self-training \cite{xie2020self}, GDA bridges large source–target gaps more effectively and has sparked both theoretical \cite{wang2022understanding} and algorithmic \cite{chen2021gradual, abnar2021gradual} advancements. Key strategies involve generating intermediate distributions via gradient flow-based geodesic paths \cite{zhuanggradual}, style-transfer interpolation \cite{marsden2024introducing}, and optimal transport \cite{he2023gradual}, while complementary methods leverage normalizing flows \cite{sagawa2022gradual}, source–target data ratio adjustments \cite{zhang2021gradual}, domain sequence discovery \cite{chen2021gradual}, and adversarial self-training \cite{shi2024adversarial} to enhance robustness and generalization.


\section{Problem Setup}

\paragraph{Domain Space}

Let $\mathcal{Z} = \mathcal{X} \times \mathcal{Y}$ be the measurable instance space, in which $\mathcal{X} \subseteq \mathbb{R}^d$ is the input space in $d$-dimensional space and $\mathcal{Y} = \{1, 2, \ldots, k\}$ is the label space, where $k$ is the number of classes.

\paragraph{Gradually shifting Domain}

In the gradually domain setting\cite{kumar2020understanding}, we have {n+1} domains indexed by \(0, 1, \ldots, n\), where the domain 0 is the source domain, domain n is the target dommain and the intermediate domains are indexed by \(1, 2, \ldots, n-1\). Those domains distribute over the instance space \(\mathcal{Z}\), donoted as \(\mathcal{D}_0, \mathcal{D}_1, \ldots, \mathcal{D}_{n} \).

\paragraph{Classification and Loss} 

The goal of classification is to learn a model \( C : \mathcal{X} \to \mathcal{Y} \) that maps input features \( x \) from the training data set \( \mathcal{D} = \{(x, y)\} \) to their corresponding labels \( y \). Considering the loss function $l$, the classifier benefit on \(\mathcal{D}_t\) is denoted by \(C\), defined as:

\vspace{-10pt}

\begin{equation}
    {C} = \arg\min_{C} \mathbb{E}_{(x, y) \sim \mathcal{D}_t} [l(C(x), y)].
\end{equation}

\vspace{-7pt}

\paragraph{Domain Adaptation} In domain adaptation settings, there are two different domains involved: the source domain \( \mathcal{D}_{\mathcal{S}} \) and the target domain \( \mathcal{D}_{\mathcal{T}} \). The model is trained on data from the source domain but is expected to generalize to the target domain, where the distribution of data may be different. The goal is to improve the model's performance on the target domain despite the lack of labeled data in the target domain, using the knowledge learned from the source domain.

\paragraph{Gradual Domain Adaptation} 

The goal of gradual domain adaptation is to learn a classifier \( C \) that generalizes well to the target domain \( \mathcal{D}_n \) by progressively transferring knowledge from the labeled source domain \( \mathcal{D}_0 \) and a series of unlabeled intermediate domains \( \mathcal{D}_1, \mathcal{D}_2, \dots, \mathcal{D}_{n-1} \). The adaptation process involves multi-step pseudo-labeling and self-training, where the model \( C_{0} \) is trained on the source domain and then adapted to the intermediate domains by the following self-training procedure \( \text{ST}(C_t, \mathcal{D}_t) \):

\vspace{-10pt}

\begin{equation}
    \text{ST}(C_t, \mathcal{D}_t) = \arg\min_{C'} \mathbb{E}_{x \sim \mathcal{D}_t} [l(C'(x), \hat{y}_t(x))].
\end{equation}

\vspace{-7pt}

In particular, \( \hat{y}_t(x) = \text{sign}(C_{t}(x)) \) is the pseudo-label generated by the model \( C_{t} \) for unlabeled data of \( \mathcal{D}_t \), where \( \mathcal{D}_t \) denotes the unlabeled intermediate domain. Meanwhile, \(C'\) is the next learned model, also denoted by \(C_{t+1}\).

\section{Methodology}

This work proposes Gradual Domain Osmosis (GDO), which reconstructs traditional self-training paradigms through a triple adaptive mechanism. The framework of GDO is shown in Figure \ref{lamda} Consider the following optimization problem:

\begin{figure}
    \centering
    \includegraphics[width=0.8\linewidth]{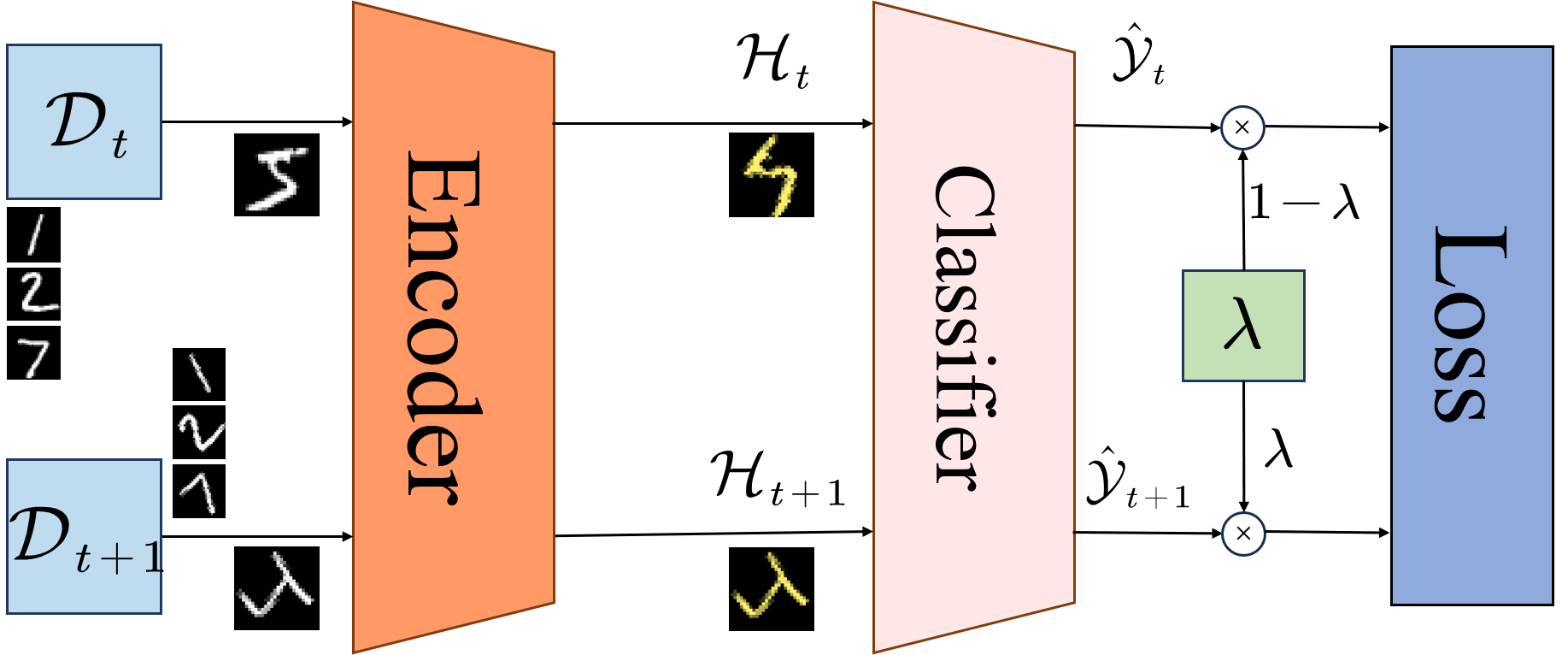}
    \caption{The framework of our GDO. It is desiqned to adapt a model from a source
domain to a target domain in a smooth and controlled manner. The hyperparameter $\lambda$ controls the trade-off between the two domains: when $\lambda=0$, the model focuses entirely on the source domain, and as $\lambda$ increases to 1, the model progressively shifts its focus to the target domain. Pseudo-labels, generated by the model itself, guide the learning process, ensuring gradual and stable adaptation. }
    \label{lamda}
\end{figure}

\begin{equation}
    \begin{gathered}
        \min_{C}\ (1-\lambda)\mathbb{E}_{x \sim \mathcal{D}_i} [l(C(x), \text{sign}(C(x)))] \\
        + \lambda \mathbb{E}_{x \sim \mathcal{D}_{i+1}} [l(C(x), \text{sign}(C(x)))].
    \end{gathered}
\end{equation}

Here \( \lambda \) is a hyperparameter that balances the trade-off between the source and target domains. With an increasing \( \lambda \) from 0 to 1, the knowledge gradually transfers from the source domain to the target domain.  For a domain sequence $\{\mathcal{D}_t\}_{t=0^{\prime}}^n$ where each domain contains $m$ data batches $\{B_t,1,...,B_{t,m}\}$, we define a time-varying classifie $C^{(t,k)}$ representing the model after the $k$-th batch update in domain $t.$ Its evolution follows:

\begin{equation}
    C^{(t,k+1)}=\Phi\left(C^{(t,k)},B_{t,k},B_{t,k+1}\right)
\end{equation}

where $\Phi$ is an incremental optimization operator based on the following objective:

\begin{equation}
    \begin{aligned}\Phi&=\arg\min_{C^{\prime}}\underbrace{\mathbb{E}_{x\sim B_{t,k}}\left[\ell_{\mathrm{ce}}(C^{\prime}(x),\hat{y}^{(t,k)}(x))\right]}_{\text{Pseudo-Label Consistency}}\\&+\alpha\underbrace{\mathbb{E}_{x\sim B_{t,k+1}}\left[\ell_{\mathrm{margin}}(C^{\prime}(x))\right]}_{\text{Prospective Margin Maximization}}\\&+\beta\underbrace{D_{\mathrm{KL}}(p(C^{\prime})\|p(C^{(t,k)}))}_{\text{Knowledge Distillation Constraint}}\end{aligned}
\end{equation}

where  dynamically generated hard pseudo-labels are represented as $\hat{y}^{(t,k)}(x)=\arg\max_yC^{(t,k)}(x)_y$, indicating the most likely class label for sample $x$ at stage $t$ and batch $k$ according to the current classifier $C^{(t,k)}$. The margin loss is defined as $\ell_{\mathrm{margin}}(z)=\max(0,1-(\max_jz_j-\max_{j\neq\hat{y}}z_j))$, where $z$ represents the unnormalized scores (logits) output by the model. 

To achieve cross-batch and cross-domain adaptation, we design a dual-timescale update rule:

\begin{equation}
    \begin{cases}\theta^{(t,k+1)}=\theta^{(t,k)}-\eta_t\nabla_\theta\mathcal{L}_\text{intra}(\theta^{(t,k)},B_{t,k})\\\phi^{(t+1,0)}=\phi^{(t,m)}-\zeta\nabla_\phi\mathcal{L}_\text{inter}(\phi^{(t,m)},\mathcal{D}_t,\mathcal{D}_{t+1})\end{cases}
\end{equation}

where $\theta$ is the feature extractor parameters updated via high-frequency batch-wise adaptation. $\phi$ is classifier head parameters updated via low-frequency domain transition.
$\mathcal{L} _{\mathrm{intra}}= \ell _{\mathrm{ce}}+ \alpha \ell _{\mathrm{margin}}$  drives rapid intra-domain adaptation.
$\mathcal{L} _{\mathrm{inter}}= \mathbb{E} _{x\sim \mathcal{D} _t\cup \mathcal{D} _{t+ 1}}[ \| \nabla _xC( x) \| _2^2]$ is the gradient alignment loss for smooth domain transitions.

\begin{figure}[t]
    \centering
    \includegraphics[width=.9\columnwidth, trim=8cm 6cm 8cm 6cm, clip, page=2]{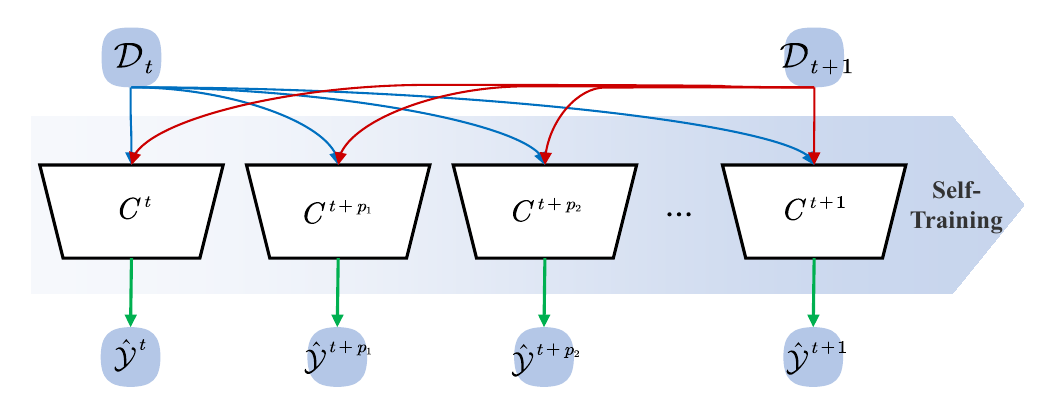}
    \caption{The process of Gradual Domain Adaptation with Dynamic Pseudo-Labeling and Self-Training. The diagram demonstrates the generation of hard pseudo-labels at each stage of the domain adaptation process. The classifiers \( C^{(t,k)} \) are progressively updated, using a dual-timescale update rule for intra-domain and inter-domain adaptation. The model updates the feature extractor parameters \( \theta \) in a high-frequency manner and classifier head parameters \( \phi \) in a low-frequency manner to facilitate smooth domain transitions. This approach promotes rapid domain adaptation and minimizes the loss between predictions and pseudo-labels across the domains \( \mathcal{D}_t \) and \( \mathcal{D}_{t+1} \).}
    \label{fig:gdo_ar}
    \label{fig:four-gda:subfig4}
\end{figure}


As shown in Fig. \ref{fig:gdo_ar}, the gradual classifier \( C \) is learned by minimizing the objective function above. The pseudo-labels are generated by the model itself, and the model is updated by minimizing the loss between the model's prediction and the pseudo-labels. \( C \) can be progressively updated by the following equation:

\begin{equation}
    \begin{gathered}
        C^{(i+\lambda)} =  \text{ST}(C,\lambda,\mathcal{D}_{i},\mathcal{D}_{i+1}) = \arg\min_{C'}\\
        (1-\lambda)\mathbb{E}_{x \sim \mathcal{D}_i} [l(C'(x), \text{sign}(C(x)))] \\
        + \lambda \mathbb{E}_{x \sim \mathcal{D}_{i+1}} [l(C'(x), \text{sign}(C(x)))].
    \end{gathered}
\end{equation}

Therefore, The dual-timescale optimization framework is implemented as algorithm \ref{gsoal}:

\begin{algorithm}[H]
\caption{Gradual Domain Osmosis (GDO)}
\label{gsoal}
\begin{algorithmic}[1]
\INPUT Source batches $\{B_{0,k}\}_{k=1}^m$, domain sequence $\{\mathcal{D}_t\}_{t=1}^n$, initial model $C^{(0,0)}$
\OUTPUT Adapted model $C^{(n,m)}$
\FOR{$t=0$ TO $n-1$}
    \FOR{$k=0$ TO $m-1$}
        \STATE Fetch batch $B_{t,k}$, generate pseudo-labels: $\hat{y}^{(t,k)}=\arg\max_yC^{(t,k)}(x)_y$
        \STATE Compute intra-domain loss: $\mathcal{L}_\mathrm{intra}=\ell_\mathrm{ce}+\alpha\ell_\mathrm{margin}$
        \STATE Update feature extractor: $\theta^{(t,k+1)}\leftarrow\theta^{(t,k)}-\eta_t\nabla_\theta\mathcal{L}_\mathrm{intra}$
        \STATE Warm-up next batch: Compute gradient alignment $\mathcal{L}_\mathrm{inter}$ for $B_{t,k+1}$
    \ENDFOR
    \STATE Inter-domain transfer: $\phi ^{( t+ 1, 0) }\leftarrow \phi ^{( t, m) }- \zeta \nabla _\phi \mathbb{E} [ \mathcal{L} _\mathrm{inter}]$
\ENDFOR
\end{algorithmic}
\end{algorithm}

\section{Theoretical Arguments}
\subsection{Error Propagation Dynamics} 

Let $\mathcal{H}$ be the hypothesis class with VC-dimension $d_{\mathcal{H}}.$ For any batch $B_t,k$,we define the empirical risk $\hat{R}_t^{(k)}(C)$ as follows:

\begin{equation}
    \hat{R}_t^{(k)}(C)=\displaystyle\frac1{|B_{t,k}|}\sum_{x\in B_{t,k}}\ell(C(x),\hat{y}^{(t,k)}(x))
\end{equation}

where $\hat{y}^{(t,k)}(x)$ represents the pseudo-label for the input $x$,and $\ell$ is the loss function. The
population risk $R_t(C)$ is defined as the expected loss over the distribution $\mathcal{D}_t:$

\begin{equation}
    R_t(C)=\mathbb{E}_{x\sim\mathcal{D}_t[\ell(C(x),y(x))]}
\end{equation}

\paragraph{Lemma 1 (Pseudo-Label Consistency)}

We want to bound the difference between the empirical risk $\hat{R}_t^{(k)}(C)$ and the population risk
$R_{t}(C).$ For $\rho(B_{t,k},B_{t,k+1})\leq\epsilon$, the pseudo-label discrepancy satisfies:

\begin{equation}
    |\hat{R}_t^{(k)}(C)-R_t(C)|\leq\sqrt{\displaystyle\frac{2d_H\log(2em/d_H)}{m}}+\epsilon L_\ell R
\end{equation}

\paragraph{Proof} 

We apply Rademacher complexity bounds with distribution shift:

\begin{equation}
    \mathbb{E}[\sup\limits_{C\in\mathcal{H}}|\hat{R}_t^{(k)}(C)-R_t(C)|]\leq2\mathfrak{R}_m(\mathcal{H})+L_\ell W_1(B_{t,k},\mathcal{D}_t)
\end{equation}

Here, $\Re_m(\mathcal{H})$ is the Rademacher complexity of the hypothesis class $\mathcal{H}$, and $W_1(B_{t,k},\mathcal{D}_t)$ is the Wasserstein distance between the empirical distribution over the batch $B_{t,k}$ and the true data distribution $\mathcal{D}_t.$ Using the fact that$W_1(B_{t,k},\mathcal{D}_t)\leq\epsilon R$ and the bound for $\Re_m(\mathcal{H})$,we get the desired result:

\begin{equation}
    |\hat{R}_t^{(k)}(C)-R_t(C)|\leq\sqrt{\displaystyle\frac{2d_{\mathcal{H}}\log(2em/d_{\mathcal{H}})}{m}}+\epsilon L_\ell R
\end{equation}

\subsection{Lyapunov Stability Analysis} 

Next, we define the Lyapunov function $V(t,k)$, which incorporates both the error and the
parameter drift:

\begin{equation}
    V(t,k)=\mathrm{Err}_{t,k}+\lambda\|\theta^{(t,k)}-\theta^{(t,k-1)}\|^2
\end{equation}

where $\mathrm{Err}_t, k= \mathbb{E} _{x\sim \mathcal{D} _t}[ \mathbb{I} ( C^{( t, k) }( x) \neq y( x) ) ]$ is the error at time $t$ and iteration $k$,and $\lambda$ is a
constant.

\paragraph{Lemma 2 (Lyapunov Drift)}

Under gradient-based updates, the Lyapunov function satisfies:

\begin{equation}
    V(t,k+1)\leq(1-\eta_t\mu)V(t,k)+\eta_t^2\sigma^2+c\epsilon\sqrt{\displaystyle\frac{\log m}{m}}
\end{equation}

where $\mu$ is the strong convexity parameter, $\sigma ^{2}$ bounds the gradient variance, $c\epsilon \sqrt {\displaystyle\frac {\log m}m}$ accounts for the discrepancy between batches.

\paragraph{Proof Sketch} 

To compute the drift $\Delta V=V(t,k+1)-V(t,k)$,we break it into two terms: the
optimization term and the parameter drift term.

\paragraph{Optimization Term:} $\mathrm{Err}_{t,k+1}-\mathrm{Err}_{t,k}$

By applying SGD convergence analysis, we get:

\begin{equation}
    \mathrm{Err}_{t,k+1}\leq\mathrm{Err}_{t,k}-\eta_t\mu\|\nabla\mathrm{Err}_{t,k}\|^2+\eta_t^2\sigma^2
\end{equation}

\paragraph{Parameter Drift Term:}$\|\theta^{(t,k+1)}-\theta^{(t,k)}\|^2$

Using the update rule $\theta^{(t,k+1)}=\theta^{(t,k)}-\eta_tg_t$, we get:

\begin{equation}
    \begin{aligned}\|\theta^{(t,k+1)}-\theta^{(t,k)}\|^2=\eta_t^2\|g_t\|^2\leq\eta_t^2G^2\end{aligned}
\end{equation}

Combining these terms, we get the Lyapunov drift bound:

\begin{equation}
    V(t,k+1)\leq(1-\eta_t\mu)V(t,k)+\eta_t^2\sigma^2+c\epsilon\sqrt{\displaystyle\frac{\log m}{m}}
\end{equation}

To bound the total error over $T$ domains, we telescope the Lyapunov inequality over the $T$
domains and $m$ batches:

\begin{equation}
    V(T,0)\leq V(0,0)\prod_{t=1}^T(1-\eta_t\mu)+\sum_{t=1}^T\left(\eta_t^2\sigma^2+c\epsilon\sqrt{\displaystyle\frac{\log m}{m}}\right)
\end{equation}

Substitute the geometric series summation:

\begin{equation}
    \prod_{t=1}^T(1-\eta_t\mu)\leq e^{-\mu\sum_{t=1}^T\eta_t}\leq e^{-\kappa\gamma_0T}
\end{equation}

where $\kappa=\displaystyle\frac\mu\gamma_02$ and $\eta_t=\displaystyle\frac{\gamma_0}{1+\epsilon t}.$ The second term accumulates as:

\begin{equation}
\begin{aligned}
    \sum_{t=1}^T\eta_t^2\sigma^2
    &\leq\sigma^2\gamma_0^2\sum_{t=1}^T\displaystyle\frac1{(1+\epsilon t)^2}\\
    &\leq\displaystyle\frac{c_1\epsilon}{\gamma_0}\sqrt{\displaystyle\frac Tm}
\end{aligned}
\end{equation}

The qeneralization error term follows from Lemma 1:

\begin{equation}
    \sum_{t=1}^Tc\epsilon\sqrt{\displaystyle\frac{\log m}m}\leq c_2\sqrt{\displaystyle\frac{\log(mT/\delta)}m}
\end{equation}

Combining all components yields the final error bound:

\begin{equation}
    \mathrm{Err}_T\le\mathrm{Err}_0\cdot e^{-\kappa\gamma_0T}+\displaystyle\frac{c_1\epsilon}{\gamma_0}\sqrt{\displaystyle\frac Tm}+c_2\sqrt{\displaystyle\frac{\log(mT/\delta)}m}
\end{equation}

where $\kappa = \displaystyle\frac {\mu \gamma _0}2{ }$ reflects the interaction between optimization rate and initial margin. $c_{1}= \displaystyle\frac {\sigma ^{2}\gamma _{0}^{2}}{\sqrt {2}}$ captures the variance-margin tradeoff. $c_2= 2c\sqrt {\log ( 1/ \delta ) }$ depends on distribution shift and confidence level.

\section{Experiments}

\subsection{Envirments Setup}

\begin{table*}[h]
\centering
\caption{Benchmarks Comparison on different datasets, including UDA methods and GDA methods.}
\resizebox{1.6\columnwidth}{!}
{%
\begin{tabular}{lccccccccccc}
\toprule UDA/GDA methods
& {Rotated MNIST} & {Color-Shift MNIST} & Portraits & Cover Type \\
\midrule
{DANN \cite{ganin2016domainneural}}
& $44.23$ & $56.5$ & $73.8$ & -
\\
{DeepCoral \cite{sun2016deep}}
& $49.6$ & $63.5$ & $71.9$ & -
\\
{DeepJDOT \cite{damodaran2018deepjdot}}
& $51.6$ & $65.8$ & $72.5$ & -
\\
\midrule
GST \cite{kumar2020understanding}
& $83.8$ & $74.0$ & $82.6$ & $73.5$
\\
IDOL \cite{chen2021gradual}
& $87.5$ & - & $85.5$ & - 
\\
GOAT \cite{he2023gradual}
& $86.4$ & $91.8$ & $83.6$ & $69.9$
\\
GGF \cite{zhuanggradual}
& $67.72$ & - & $86.16$ & -
\\
CNF \cite{sagawa2025gradual}
& $62.55$ & - & $84.57$ & - 
\\
GDO (Ours)
& $\mathbf{97.6}$ & $\mathbf{98.3}$ & $\mathbf{86.1}$ & $\mathbf{74.2}$
\\
\bottomrule
\end{tabular}%
}
\label{tab:comp-uda}
\end{table*}

To empirically validate our method, we examine gradual self-training on two synthetic and two real datasets, including Rotated MNIST, Color-Shift MNIST, Portraits Dataset \cite{ginosar2015century} and Cover Type Dataset\cite{covertype_31}.

\paragraph{Rotated MNIST}
This is a semi-synthetic dataset based on the widely used MNIST dataset\cite{deng2012mnist}. Following \citet{he2023gradual}, it consists of 50,000 images for the source domain (original MNIST images) and the same 50,000 images rotated by 45 degrees to form the target domain. The intermediate domains are evenly distributed between the source and target domains, with rotations gradually changing from 0 to 45 degrees.

\paragraph{Color-Shift MNIST}
This dataset is built by applying a color shift to the MNIST dataset. Following \citet{he2023gradual}, it consists of the same 50,000 images as the source domain, in which the pixel values are normalized to $[0,1]$, with the pixel values shifted to be in the range $[1, 2]$ for the target domain. The intermediate domains are evenly distributed between the source and target domains, resulting in various color shifts.

\paragraph{Portraits Dataset \cite{ginosar2015century}}
This dataset contains portraits of high school seniors from 1905 to 2013, primarily used for gender classification tasks. Following \citet{kumar2020understanding}, the dataset is arranged chronologically, from front to back, with every 2000 images set as a domain, and a total of 9 domains are set, containing the first being the source domain, the middle seven intermediate domains, and the last being the target domain. All data is processed to 32 x 32 pixel size.

\paragraph{Cover Type Dataset\cite{covertype_31}}
The Cover Type dataset is a tabular dataset designed to predict the type of forest cover and contains 54 features. Following \citet{kumar2020understanding}, we focus on the first two categories of forest cover: spruce fir and Rocky Mountain pine. We sort the dataset based on the horizontal distance from the nearest water body. The first 50,000 data are used as the source domain, followed by every 40,000 subsequent data, creating a total of ten intermediate domains, and the last 50,000 data are treated as the target domain.

\paragraph{Implementation} For the Rotated MNIST, Color-Shift MNIST, and Portraits datasets, a convolutional neural network (CNN) architecture was implemented consisting of three convolutional layers, each with 32 channels. The encoder is followed by a fully connected classifier with two hidden layers, each containing 256 units. For the Cover Type dataset, a similar architecture was employed, utilizing three fully connected layers with ReLU activations, where the sizes of the hidden layers increase progressively from 128 to 256 to 512 units. The final layer corresponds to the number of classes in the dataset. Optimization was performed using the Adam optimizer \cite{kingma2014adam}, while regularization was achieved via Dropout \cite{srivastava2014dropout}. Batch Normalization \cite{ioffe2015batch} was applied to enhance the stability of training. The transport network architecture integrates generators, which are constructed using a residual block containing three linear layers. The discriminator comprises three linear layers, each with 128 hidden units, and employs ReLU activations. The number of intermediate domains generated between the source and target domains was treated as a hyperparameter, with model performance evaluated for 0, 1, 2, 3, or 4 intermediate domains. All experiments were conducted on NVIDIA RTX 4090 GPUs.

\paragraph{Benchmarks} To verify the effectiveness of our GDO, we compare it with serval state-of-the-art (SOTA) methods, including 3 UDA methods, DANN \cite{ganin2016domain}, DeepCoral \cite{sun2016deep}, DeepJDOT \cite{damodaran2018deepjdot} and 5 GDA methods, GST \cite{kumar2020understanding}, IDOL \cite{chen2021gradual}, GOAT \cite{he2023gradual}, GGF \cite{zhuanggradual}, and GNF \cite{sagawa2025gradual} using the same training datasets with ours.

\subsection{Results}

\begin{table}[h!]
\centering
\caption{Comparison of the accuracy of our method for different given intermediate domains (including source and target domains) on the \textbf{Rotated MNIST}  dataset, as well as the 95\% confidence interval of the mean across 5 runs.}
\resizebox{\columnwidth}{!}{%
\begin{tabular}{cccccc}
\toprule
\multicolumn{2}{l}{\# Given} & \multicolumn{3}{c}{\# Inter-domain counts in GDO} \\
\small Domains & 0  & 1 & 2 & 3 & 4 \\
\midrule
2 & $81.5\pm 1.3$ & $82.8\pm 3.5$ & $81.7\pm 4.4$ & $82.0\pm 6.4$ & $\mathbf{83.6\pm 1.2}$ \\
3 & $95.9\pm 0.4$ & $96.5\pm 0.8$ & $96.8\pm 0.3$ & $\mathbf{96.9\pm 0.1}$ & $96.8\pm 0.1$ \\
4 & $96.5\pm 0.2$ & $\mathbf{96.9\pm 0.1}$ & $96.8\pm 0.0$ & $96.8\pm 0.0$ & $96.8\pm 0.1$ \\
5 & $96.8\pm 0.1$ & $96.9\pm 0.1$ & $96.7\pm 0.1$ & $96.9\pm 0.0$ & $\mathbf{97.0\pm 0.1}$ \\
6 & $96.9\pm 0.1$ & $96.9\pm 0.1$ & $96.9\pm 0.1$ & $\mathbf{97.6\pm 0.0}$ & $96.9\pm 0.0$ \\
\bottomrule
\end{tabular}%
}
\label{tab:ans1}
\end{table}

\begin{table}[h!]
\centering
\caption{Comparison of the accuracy of our method for different given intermediate domains (including source and target domains) on the \textbf{Color-Shift MNIST}  dataset, as well as the 95\% confidence interval of the mean across 5 runs.}
\resizebox{\columnwidth}{!}{%
\begin{tabular}{cccccc}
\toprule
\multicolumn{2}{l}{\# Given} & \multicolumn{3}{c}{\# Inter-domain counts in GDO} \\
\small Domains & 0  & 1 & 2 & 3 & 4 \\
\midrule
2 & $86.2\pm 0.1$ & $96.5\pm 0.0$ & $\mathbf{87.2\pm 3.6}$ & $86.6\pm 0.0$ & $86.6\pm 0.1$ \\
3 & $98.1\pm 0.0$ & $98.2\pm 0.0$ & $98.2\pm 0.0$ & $\mathbf{98.3\pm 0.0}$ & $\mathbf{98.3\pm 0.0}$ \\
4 & $\mathbf{98.3\pm 0.0}$ & $\mathbf{98.3\pm 0.0}$ & $\mathbf{98.3\pm 0.0}$ & $\mathbf{98.3\pm 0.0}$ & $\mathbf{98.3\pm 0.0}$ \\
5 & $\mathbf{98.3\pm 0.0}$ & $\mathbf{98.3\pm 0.0}$ & $\mathbf{98.3\pm 0.0}$ & $98.2\pm 0.0$ & $\mathbf{98.3\pm 0.0}$ \\
6 & $\mathbf{\mathbf{98.3\pm 0.0}}$ & $\mathbf{98.3\pm 0.0}$ & $\mathbf{98.3\pm 0.0}$ & $98.2\pm 0.0$ & $98.2\pm 0.0$ \\
\bottomrule
\end{tabular}%
}
\label{tab:ans2}
\end{table}

\begin{table}[h!]
\centering
\caption{Comparison of the accuracy of our method for different given intermediate domains (including source and target domains) on the \textbf{Portraits}  dataset, as well as the 95\% confidence interval of the mean across 5 runs.}
\resizebox{\columnwidth}{!}{%
\begin{tabular}{cccccc}
\toprule
\multicolumn{2}{l}{\# Given} & \multicolumn{3}{c}{\# Inter-domain counts in GDO} \\
\small Domains & 0  & 1 & 2 & 3 & 4 \\
\midrule
2 & $83.7\pm 0.3$ & $84.4\pm 0.8$ & $\mathbf{85.1\pm 0.6}$ & $\mathbf{85.1\pm 0.8}$ & $\mathbf{85.1\pm 0.5}$ \\
3 & $84.1\pm 0.3$ & $84.6\pm 0.2$ & $\mathbf{84.8\pm 0.5}$ & $84.7\pm 0.1$ & $84.7\pm 0.2$ \\
4 & $83.8\pm 0.7$ & $\mathbf{84.0\pm 0.1}$ & $\mathbf{84.0\pm 0.2}$ & $83.8\pm 0.3$ & $83.9\pm 0.1$ \\
5 & $84.8\pm 0.4$ & $84.8\pm 0.4$ & $84.8\pm 0.4$ & $84.8\pm 0.2$ & $\mathbf{84.9\pm 0.4}$ \\
6 & $85.3\pm 0.9$ & $\mathbf{86.1\pm 0.4}$ & $85.8\pm 1.0$ & $85.6\pm 1.8$ & $85.4\pm 0.9$ \\
\bottomrule
\end{tabular}%
}
\label{tab:ans3}
\end{table}

\begin{table}[h!]
\centering
\caption{Comparison of the accuracy of our method for different given intermediate domains (including source and target domains) on the \textbf{Cover Type}  dataset, as well as the 95\% confidence interval of the mean across 5 runs.}
\resizebox{\columnwidth}{!}{%
\begin{tabular}{cccccc}
\toprule
\multicolumn{2}{l}{\# Given} & \multicolumn{3}{c}{\# Inter-domain counts in GDO} \\
\small Domains & 0  & 1 & 2 & 3 & 4 \\
\midrule
2 & $69.7\pm 0.0$ & $70.0\pm 0.0$ & $70.5\pm 0.1$ & $71.2\pm 0.0$ & $\mathbf{71.8\pm 0.1}$ \\
3 & $70.1\pm 0.0$ & $72.3\pm 0.0$ & $73.9\pm 0.0$ & $74.3\pm 0.0$ & $\mathbf{74.4\pm 0.1}$ \\
4 & $71.5\pm 0.0$ & $73.8\pm 0.0$ & $\mathbf{74.2\pm 0.0}$ & $74.0\pm 0.0$ & $\mathbf{74.2\pm 0.0}$ \\
5 & $72.5\pm 0.1$ & $\mathbf{74.2\pm 0.0}$ & $74.1\pm 0.0$ & $\mathbf{74.2\pm 0.1}$& $73.9\pm 0.0$ \\
6 & $73.1\pm 0.0$ & $\mathbf{74.1\pm 0.0}$ & $73.4\pm 0.0$ & $73.9\pm 0.0$ & $73.4\pm 0.0$ \\
\bottomrule
\end{tabular}%
}
\label{tab:ans4}
\end{table}

\begin{figure*}[htbp!]
    \centering
    \subfigure[Rotated MNIST]{\includegraphics[height = 4.2cm, trim=0cm 0cm 0cm 0cm, clip]{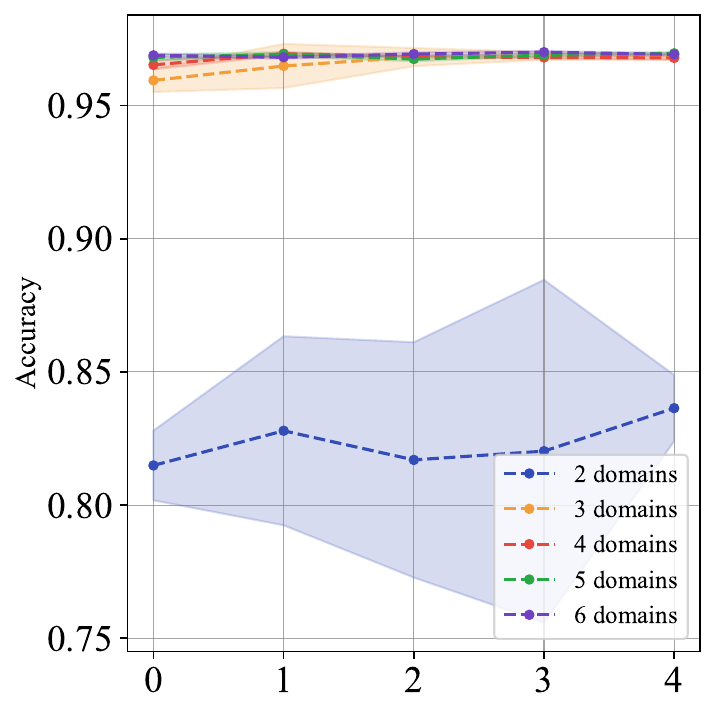}\label{fig:abl-sub1}}
    \hfill
    \subfigure[Color-Shift MNIST]{\includegraphics[height = 4.2cm, trim=0cm 0cm 0cm 0cm, clip]{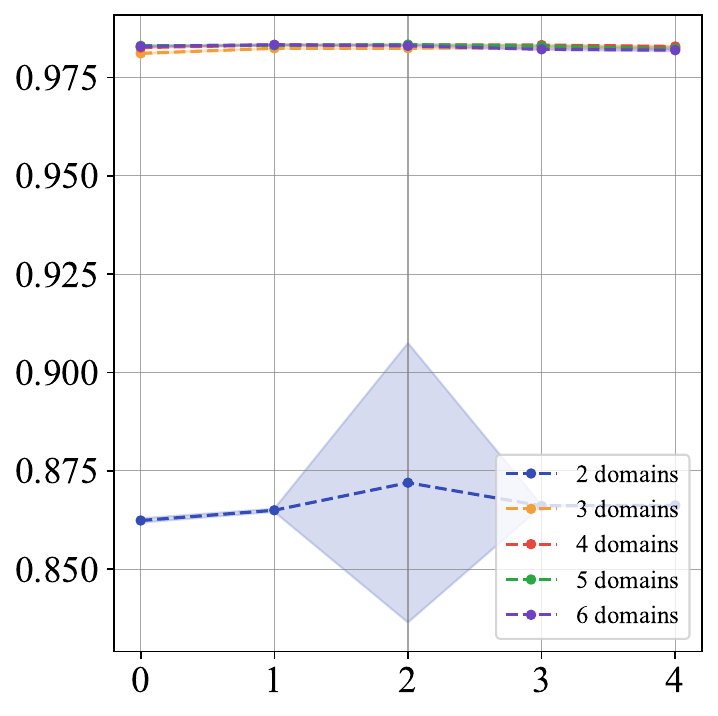}\label{fig:abl-sub2}}
    \hfill
    \subfigure[Portraits]{\includegraphics[height = 4.2cm, trim=0cm 0cm 0cm 0cm, clip]{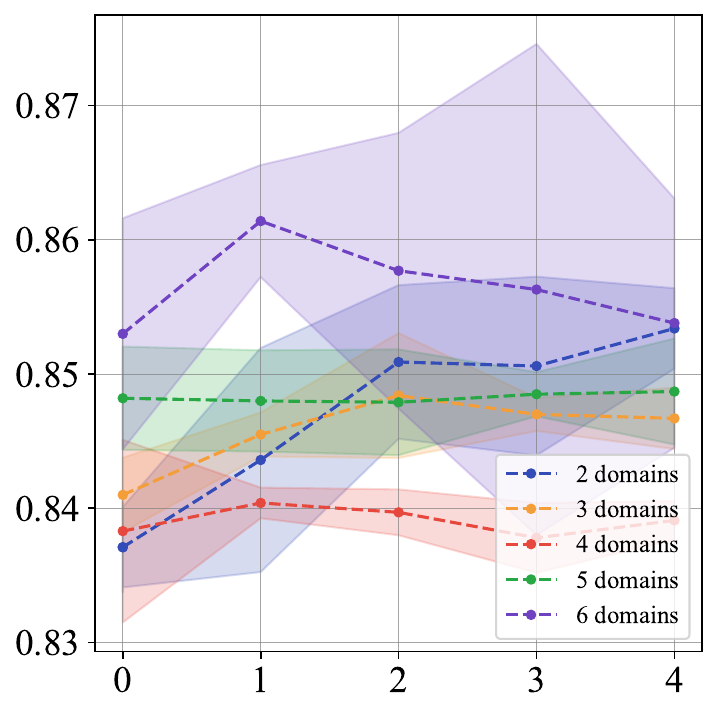}\label{fig:abl-sub3}}
    \hfill
    \subfigure[Covertype]{\includegraphics[height = 4.2cm, trim=0cm 0cm 0cm 0cm, clip]{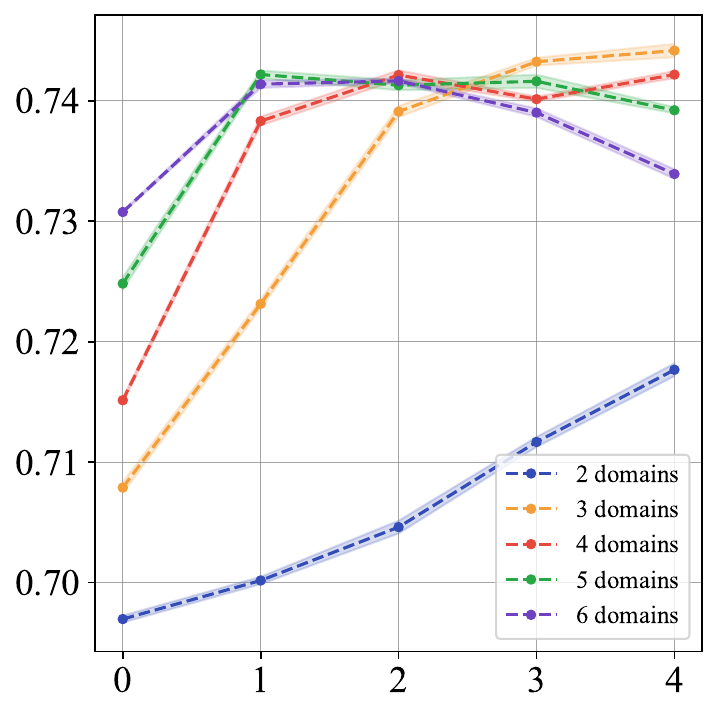}\label{fig:abl-sub4}}
    \caption{Ablation experiments with different domain conditions are conducted on four datasets. Two domains represent no intermediate domains (only source domain and target domain), and the added domains are all intermediate domains.}
    \label{fig:ablation}
\end{figure*}

Table \ref{tab:comp-uda} demonstrates our method's superior performance across all benchmark datasets compared to existing UDA and GDA approaches. Our GDO achieves state-of-the-art results with significant margins. Specifically, Our GDO achieved 11.2\%, 6.5\%, 0.94\%, and 4.3\% improvement in accuracy compared to the second-best results on Rotated MNIST, Color-Shift MNIST, Portraits, and Cover Type datasets. The results validate that our GDO better handles continuous distribution shifts compared to static domain alignment approaches like DANN or single-step gradual methods.

We present a comparative analysis of our proposed SWAT method across multiple datasets, namely Rotated MNIST, Color-Shift MNIST, Portraits, and Cover Type, as detailed in Tables \ref{tab:ans1} to \ref{tab:ans4}. Each experiment was conducted multiple times, and the results are reported as mean values accompanied by their respective variance intervals. The leftmost column of each table displays the performance achieved using adversarial training alone, corresponding to the baseline approach without the incorporation of flow matching.

In Tables \ref{tab:ans1} to \ref{tab:ans4}, the column labeled ``\# Given Domains" denotes the total number of domains involved in each experiment, including both source and target domains. The columns titled ``Inter-domain counts in SWAT" represent the number of inter-domain steps performed between the specified domains within the dataset. The total number of training steps is given by the expression:$(\# \text{Given Domains} - 1) \times (\# \text{Inter-domain counts in SWAT} + 1) + 1$, which accounts for the self-training procedure involving both the GAN, the encoder \(f\), and the classifier \(g\). For example, when four domains and three intermediate steps are considered, the total number of training steps is calculated as $(4 - 1) \times (3 + 1) + 1 = 13 \text{ small steps}.$

The experimental results demonstrate the effectiveness of the SWAT method across the datasets under consideration—Rotated MNIST, Color-Shift MNIST, Portraits, and Cover Type. For each dataset, we varied the number of given domains and the number of inter-domain steps in the SWAT procedure. The performance of the model was evaluated as the number of inter-domain steps increased, providing a detailed analysis of the influence of these factors on the overall effectiveness of the domain adaptation process.
In addition, small standard deviations ($\leq$1.2\%) confirm method reliability.

\subsection{Ablation Study}

In order to verify the influence of the intermediate domain on the experimental results, we conducted experiments on four datasets without GDO, with 2 domains (i.e., only the source domain and the target domain) to 7 domains (i.e., including 5 intermediate domains). The results are shown in the figure \ref{fig:ablation}. When the intermediate domain is included, the accuracy is greatly improved in most experimental cases compared with the case without the intermediate domain, and the standard deviation is also reduced. This shows that progressive domain adaptation has a great influence on improving domain generalization.

\section{Conclustion}
In this paper, we introduced Gradual Domain Osmosis (GDO), a novel method for Gradual Domain Adaptation (GDA) that effectively addresses the challenges of smooth knowledge migration across domains. By dynamically adjusting the hyperparameter $\lambda$, which balances the weight between the source and target domains, GDO allows for a progressive, efficient transfer of knowledge through a self-training framework. This approach mitigates domain bias, ensuring that the model generalizes well in a progressively adapted intermediate domain. Our experimental evaluations on multiple datasets, demonstrate that GDO outperforms existing state-of-the-art methods in both accuracy and robustness.


\section*{Impact Statement}

This paper presents work whose goal is to advance the field of Machine Learning. There are many potential societal consequences of our work, none which we feel must be specifically highlighted here.

\bibliography{paper}
\bibliographystyle{icml2025}

\newpage
\onecolumn
\appendix

\end{document}